\documentclass{article}

\usepackage[numbers]{natbib}
\usepackage[final,nonatbib]{neurips_2020}

\usepackage[utf8]{inputenc} 
\usepackage[T1]{fontenc}    
\usepackage{hyperref}       
\usepackage{url}            
\usepackage{booktabs}       
\usepackage{amsfonts}       
\usepackage{nicefrac}       
\usepackage{microtype}      

\usepackage{amsmath}
\usepackage{amsthm}
\usepackage{bbm}
\usepackage{bbold}
\usepackage{multirow}
\usepackage{graphicx}
\usepackage{subfigure}
\usepackage{wrapfig}
\usepackage{caption}

\title{Empirical Likelihood for Contextual Bandits}

\author{%
	Nikos Karampatziakis \\
	Microsoft Dynamics 365 AI \\
	\texttt{nikosk@microsoft.com}
	\And
	John Langford \\
	Microsoft Research \\
	\texttt{jcl@microsoft.com}
	\And
	Paul Mineiro \\
	Microsoft Research \\
	\texttt{pmineiro@microsoft.com}
}

\DeclareMathOperator*{\argmax}{arg\,max}

\newtheorem{theorem}{Theorem}
\newtheorem{lemma}{Lemma}

\newcommand{\E}{\mathbb{E}}
\newcommand{\Cov}{\mathrm{Cov}}
\newcommand{\C}{\mathcal{C}}

\newcommand{\redactedurl}[1]{\url{http://redacted.for.review}}

\DeclareMathOperator{\erf}{erf}

\begin{document}

\maketitle

\newif\ifisappendix
\isappendixfalse

\begin{abstract}
	We propose an estimator and confidence interval for computing the value
	of a policy from off-policy data in the contextual bandit setting.  To
	this end we apply empirical likelihood techniques to formulate our
	estimator and confidence interval as simple convex optimization
	problems.  Using the lower bound of our confidence interval, we then
	propose an off-policy policy optimization algorithm that searches for
	policies with large reward lower bound.  We empirically find that both
	our estimator and confidence interval improve over previous proposals
	in finite sample regimes.  Finally, the policy optimization algorithm
	we propose outperforms a strong baseline system for learning from
	off-policy data.
\end{abstract}

\section{Introduction}
Contextual Bandits~\cite{EXP4,epochgreedy} are now in widespread practical use
(\cite{news, netflix, wellness}).  Key to their success is the ability to do
\emph{off-policy} or \emph{counterfactual estimation} \citep{HT52} of the value
of any policy enabling sound train/test regimes similar to supervised learning.
However, off-policy evaluation requires more data than supervised learning to
produce estimates of the same accuracy. This is because off-policy data needs
to be importance-weighted and accurate estimation for importance-weighted data
is still an active research area. How can we find a tight confidence interval
(CI) on counterfactual estimates?  And since tight CIs are deeply dependent on
the form of their estimate, how can we find a tight estimate?  And given what
we discover, how can we leverage this for improved learning algorithms?

\begin{figure*}[t]
\centering
\subfigure{
  \includegraphics[width=0.48\textwidth]{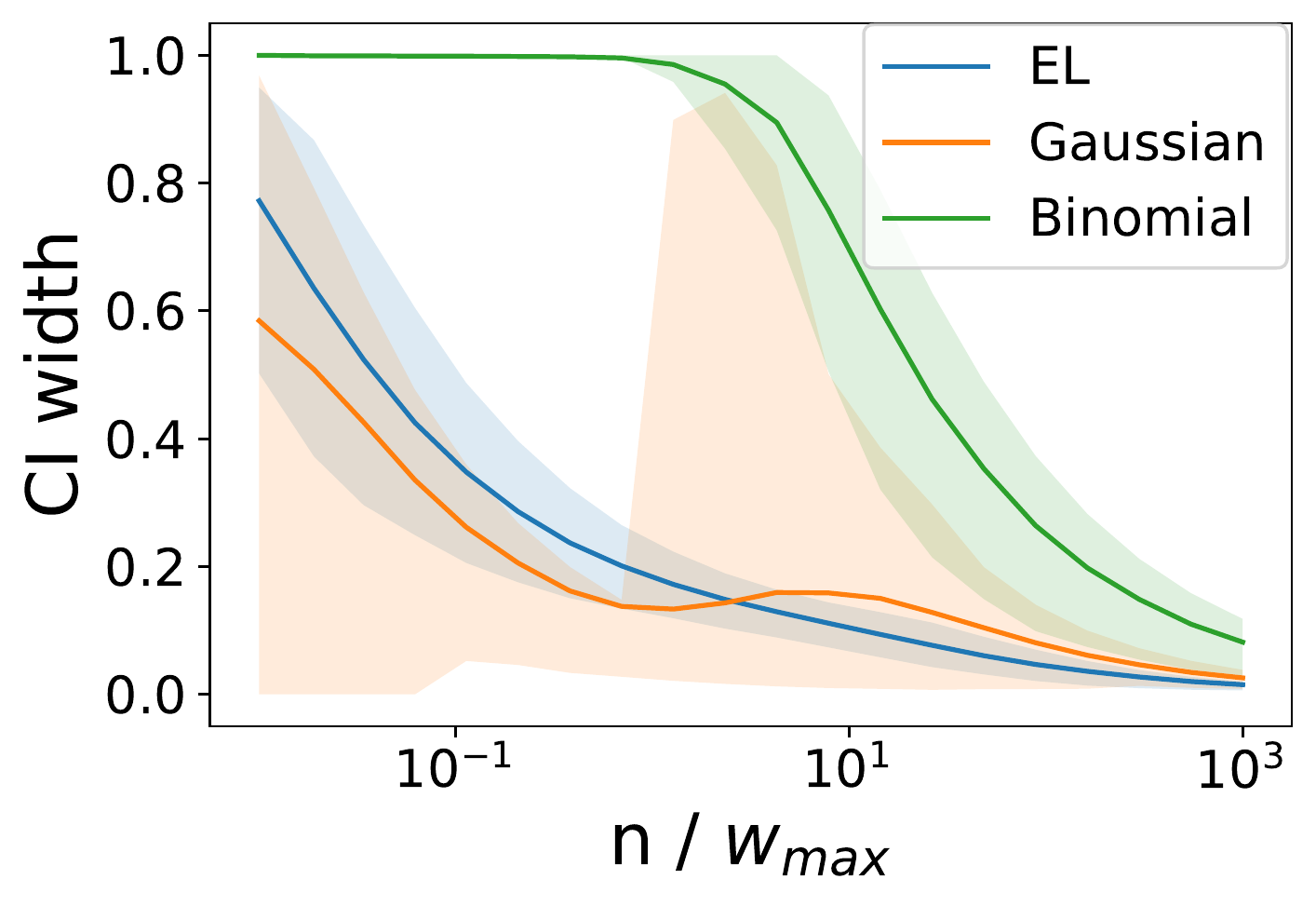}
}
\hfill
\subfigure{
  \includegraphics[width=0.48\textwidth]{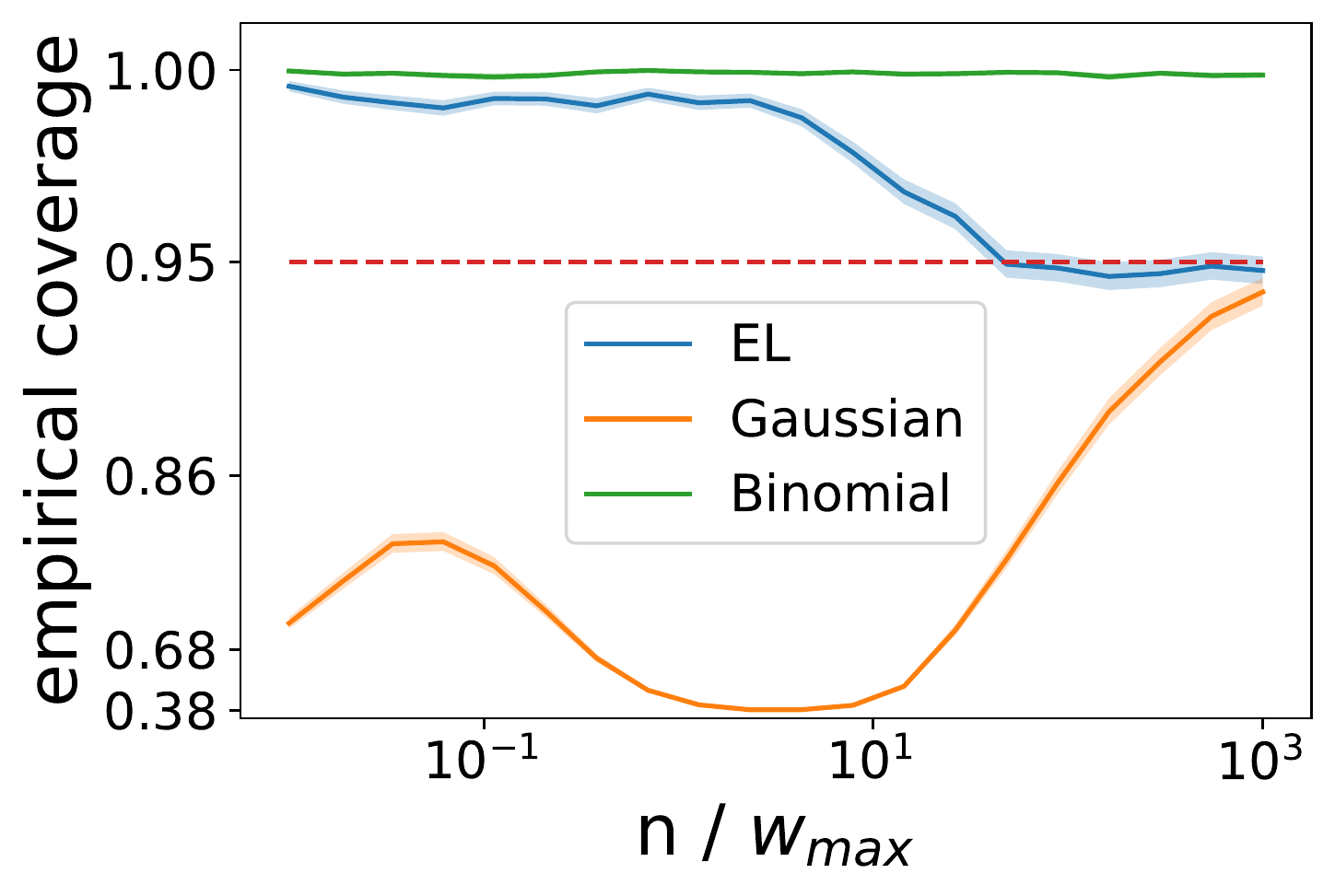}
}
\caption{A comparison of confidence intervals on contextual
bandit data.  The EL confidence interval is dramatically tighter
than an approach based on a binomial confidence interval while
avoiding chronic undercoverage as per the asymptotic Gaussian
confidence interval.  In some regimes, the asymptotic
Gaussian CI both undercovers \emph{and} has greater average width.
This is possible as the EL CI has a different functional form than
a multiplier on the Gaussian CI.  On the left, shaded area represents
90\% of the empirical distribution indicating the EL CI width varies
less over realizations.  On the right, shaded area represents 4 times the
standard error of the mean indicating coverage differences are everywhere
statistically significant.
}
\label{fig:cisynthetic}
\end{figure*}

We discover good answers to these questions through the application of
empirical likelihood \cite{owen2001empirical}, a nonparametric maximum
likelihood approach that treats the sample as a realization from a multinomial
distribution with an infinite number of categories. Like a likelihood method,
empirical likelihood (EL) adapts to the difficulty of the problem in an
automatic way and results in efficient estimators.  Unlike parametric
likelihood methods, we do not need to make any parametric assumptions about the
data generating process. We do assume that the expected importance weight is 1,
a nonparametric moment condition that is supposed to hold for correctly
collected off-policy data.  Finally, EL-based estimators and confidence
intervals can be computed by efficient algorithms that solve low dimensional
convex optimization problems. Figure~\ref{fig:cisynthetic} shows a preview of
our results. 

In section~\ref{subsec:estimation} we introduce our estimator. The estimator is
computationally tractable, requiring a bisection search over a single scalar,
has provably low bias (see Theorem~\ref{thm:finitebias}) and in section
\ref{subsec:expestimation} we experimentally demonstrate performance exceeding
that of popular alternatives. 

The estimator leads to an asymptotically exact confidence interval for
off-policy estimation which we describe in section~\ref{subsec:ci}.  Other CIs
are either narrow but fail to guarantee prescribed coverage, or guarantee
prescribed coverage but are too wide to be useful.  Our interval is narrow and
(despite having only an asymptotic guarantee) empirically approaches nominal
coverage from above as in Figure~\ref{fig:cisynthetic} and
Table~\ref{tab:ciresults}.  Finally, in section~\ref{subsec:lblearn}, we use
our CI to construct a robust counterfactual learning objective. We experiment
with this in section~\ref{subsec:explbo} and empirically outperform a strong
baseline.

We now highlight several innovations in our approach:
\begin{itemize}
  \item We use a nonparametric likelihood approach.
  This maintains \cite{kitamura2001asymptotic} some
  of the asymptotic optimality results known for likelihood 
  in the multinomial (hence well-specified) case\cite{hoeffding1965asymptotically}. 
  \item We prove a finite sample result on the bias of our
  estimator. This also implies
  our estimator is asymptotically consistent.
  \item Our CI considers a large set 
  of plausible worlds (alternative hypotheses) from which 
  the observed off-policy data could have come from. One 
  implication (cf.~section~\ref{sec:constraint}) is that 
  for binary rewards the CI lower 
  bound will be $<1$ (and $> 0$)
  even if all observed rewards are $1$.
  \item We show how to compute the confidence interval directly,
  saving a factor of $\log(1/\epsilon)$ in time complexity 
  compared to standard implementations of EL for general settings.
  \item We propose a learning objective that searches for a policy
  with the best lower bound on its reward and draw connections with
  distributionally robust optimization.
\end{itemize}

\section{Related Work}

There are many off-policy estimators for contextual bandits.  The "Inverse
Propensity Score" (IPS)~\cite{HT52} is unbiased, but has high variance.  The
Self-Normalized IPS (SNIPS)~\cite{swaminathan2015self} estimator trades off
some bias for better mean squared error (MSE).  Our estimator has bias of the
same order as SNIPS and empirically better MSE. The EMP estimator
of~\cite{kallus2019intrinsically} also uses EL techniques and we will explain
the differences in detail in section~\ref{subsec:estimation}. Critically, it
would be challenging to use EMP to construct a CI with correct coverage for
small samples, as we will explain in section~\ref{sec:constraint}.  An
orthogonal way to reduce variance is to incorporate a reward estimator as in
the doubly robust (DR) estimator and associated
variants~\cite{RnR,dudik2011doubly,SWITCH,vlassis2019design}.  The estimator
presented here is a natural alternative to IPS and SNIPS and can naturally
replace the IPS part of a doubly robust estimator.

There is less work on off-policy CIs for contextual bandits.  A simple baseline
randomly rounds the rewards to $\{0, 1\}$ and the importance weights to 0 or
the largest possible weight value and applies a Binomial confidence interval.
Another simple asymptotically motivated approach, previously applied to
contextual bandits~\cite{li2015counterfactual}, is via a Gaussian
approximation.  The EL confidence intervals are also asymptotically motivated
but empirically approach nominal coverage from above and are much tighter than
the Binomial confidence interval. In \cite{bottou2013counterfactual} empirical
Bernstein bounds or Gaussian approximations are combined with clipping of large
importance weights to trade bias for variance.  This requires hyperparameter
tuning whereas EL provides parameter-free CIs. Similar ideas to ours have been
used for the upper confidence bound in the Empirical KL-UCB
algorithm~\cite{cappe2013kullback}, an on-policy algorithm for multi-armed
bandits. As detailed in section~\ref{sec:constraint}, both constructions need
to consider some events that may not be in the data. While this happens without
explicit data augmentation, it is analogous to the use of explicitly augmented
MDPs for off-policy estimation in Markov Decision Processes\cite{liu2019off}. 

Learning algorithms for contextual bandits include theoretical~\cite{EXP4,
epochgreedy}, reduction oriented~\cite{dudik2011doubly},
optimization-based~\cite{swaminathan2015batch}, and Bayesian~\cite{BCB}
algorithms. A recent paper about empirical contextual bandit
learning~\cite{Bakeoff} informs our experiments.

Ideas from empirical likelihood have previously been applied to robust
supervised learning~\cite{duchi2016statistics}.  Our combination of CIs with
learning is a contextual bandit analogue to robust supervised learning.
Regularizing counterfactual learning via lower-bound optimization has been
previously considered, e.g., based upon empirical Bernstein
bounds~\cite{swaminathan2015batch} or divergence-based trust regions grounded
in lower bounds from conservative policy
iteration~\cite{schulman2015trust,kakade2002approximately}.

\section{Notation and Warm-up}

We consider the off-policy contextual bandit problem, with contexts
$x \in \mathcal{X}$, a finite set of actions $A$, and bounded real
rewards $\textbf{r} \in A \to [0, 1]$.  The environment generates
i.i.d. context-reward pairs $(x, \textbf{r}) \sim D$ and first reveals
$x$. Then an action $a \in A$ is sampled and the reward $\textbf{r}(a)$
is revealed.

Let $\pi$ be the policy whose value we want to estimate.  For off-policy
estimation we assume a dataset $\{ (x_n, a_n, p_n, r_n) \}_{n=1}^{N}$,
generated from an arbitrary sequence of historical stochastic policies
$h_n$, with $p_n \doteq h_n(a_n|x_n)$ and $r_n \doteq \textbf{r}_n(a_n)$.
Let $w(a)\doteq \frac{\pi(a|x)}{h(a|x)}$ be a random variable denoting
the density ratio between $\pi$ and $h$ and 
$w_n \doteq \frac{\pi(a_n|x_n)}{h_n(a_n|x_n)}$ its realization.
We assume $\pi \ll h_n$ (absolute continuity), 
and that $w \in [w_{\min}, w_{\max}]$.\footnote{ $w_{\min}=0$ is always a lower
bound, but $w_{\max}$ is application dependent. To ensure $\pi \ll h_n$ so that
estimation is consistent, it is common to enforce, for every action $a'$,
$h_n(a'|x_n)\geq p_{\min}$.  Then $w_{\max}\leq 1/{p_{\min}}$.} The value of
$\pi$ is defined as $V(\pi)= \E_{(x,\textbf{r})\sim
D,a\sim\pi(\cdot|x)}[\textbf{r}(a)]$. Since we don't have data from $\pi$, but
from $h_n$ we use importance weighting to write $V(\pi)=\E_{(x,\textbf{r})\sim
D,a\sim h(\cdot|x)}[w(a)\textbf{r}(a)]$.  The inverse propensity score (IPS)
estimator is a direct implementation of this: $V^{\textrm{IPS}}(\pi) =
\frac{1}{N}\sum_{n=1}^N w_n r_n$.  We can do better by observing that each
policy $h_n$ is created using data before time $n$.  Formally, let $\{
\mathcal{F}_n \}$ be the filtration generated by $\{ (x_k, a_k, p_k,
r_k) \}_{k < n}$, and assume $\{ h_n \}$ is $\{ \mathcal{F}_n\}$-adapted. Let
$\E_n[\cdot] \doteq \E[\cdot|\mathcal{F}_n]$.  These observations allow us to
note that $\forall n: \mathbb{E}_n[w(a)] = 1$. This moment condition has been
used for variance reduction (e.g in the SNIPS estimator). We also observe that
$
m_n(v) = \left(
\begin{matrix}
\sum_{k \leq n} \left( w_{k} r_{k} - v \right) \\
\sum_{k \leq n} \left( w_{k} - 1 \right) \\
\end{matrix}
\right)
$
is a martingale sequence when $v=V(\pi)$.  This observation will allow us to
develop consistent estimators even with a non-stationary behavior policy.

\subsection{Pedagogical Example}

Suppose $\pi$ is deterministic, $\textbf{r}$ is binary-valued, and $h_n$ is the
same $\epsilon$-greedy policy for all $n$. In this case there are only 3
possible values\footnote{$\pi(a_n|x_n) \in \{0, 1\}$ and $h_n(a_n|x_n)$ has two
possible values.} for the importance weight $w_n =
\frac{\pi(a_n|x_n)}{h_n(a_n|x_n)}$; 2 possible values for the reward, and the
data is an i.i.d. sample.  The observed data can be reduced to a histogram with
6 bins.  To construct an estimator and a confidence interval we will reason
about plausible worlds that could have generated the data. In particular each
of these worlds induces a joint distribution over importance weights and
rewards. Let $Q_{w,r}^*$ denote the true probability of $(w, r)$ under the
logging policy. Its maximum likelihood estimator is 
\begin{equation*}
\begin{split}
Q^{\textrm{mle}}=\argmax_{Q \in \Delta}\left\{\left. \sum_n \log\left(Q_{w_n,r_n}\right) \right| \E_{Q}\left[w\right] = 1 \right\},
\end{split}
\end{equation*}
where $\Delta$ is the simplex. The constraint enforces
that the counterfactual distribution under $\pi$ normalizes; we
discuss the implications in section~\ref{sec:constraint}.  Associated
with any maximizer $Q^{\textrm{mle}}$ is a corresponding value
estimate $\hat V(\pi) = \mathbb{E}_{Q^{\textrm{mle}}}\left[w r\right]$.
Absent the constraint, the maximizing $Q$ would have been the empirical 
distribution and $\hat V(\pi)$ would be $V^{\textrm{IPS}}(\pi)$. 
Furthermore, to find an asymptotic CI for $\hat V(\pi)$,
we can use Wilks' theorem. Define the maximum profile likelihood at $v$:
\begin{equation}
L(v)=\sup_{Q \in \Delta}\left\{\left. \sum_n \log\left(Q_{w_n,r_n}\right) \right| \mathbb{E}_Q[w] = 1, \mathbb{E}_Q[w r] = v \right\}.
\label{eq:primalprof}
\end{equation}
Let $Q^{\textrm{prof}}(v)$ be the maximizing $Q$ for $L(v)$.
Wilks' Theorem says that
$-2(L(V(\pi))-\sum_n\log(Q^{\textrm{mle}}_{w_n,r_n})) \to \chi_{(1)}^2$
in distribution as $n\to\infty$. Letting
$\chi_{(1)}^{2,1-\alpha}$ be the $1-\alpha$-quantile of a $\chi$-square
distribution with one degree of freedom, an asymptotic
$1-\alpha$-confidence interval is
\[
\left\{ v \biggr| \sum_n\log(Q^{\textrm{mle}}_{w_n,r_n}) - \sum_n\log(Q_{w_n,r_n}^{\textrm{prof}}(v)) \leq \frac{1}{2} \chi_{(1)}^{2,1-\alpha} \right\}.
\]
That is, if for a candidate $v$ there exists a distribution 
$Q^{\textrm{prof}}(v)$ over $(w,r)$ pairs such that 
$\mathbb{E}_{Q^{\textrm{prof}}(v)}[w]=1$,
$\mathbb{E}_{Q^{\textrm{prof}}(v)}[wr]=v$, 
and the data likelihood is high then $v$ should
be in the CI for $V(\pi)$.
\cite{hoeffding1965asymptotically} shows that for multinomials
this is the tightest $1-\alpha$-confidence interval as $n\to \infty$ and $\alpha\to 0$.

The value estimate is not necessarily unique if there are multiple
distributions $Q^{\textrm{mle}}$ which obtain the maximum, but all
value estimates are contained in the $\alpha \to 1$ limit of the above
CI.  For instance, if all observed importance weights
are zero by chance, $Q^{\textrm{mle}}$ must place some mass on a $(w, r)$
with $w > 1$ to satisfy $\E_Q[w]=1$, but the likelihood is not sensitive
to the value of $r$.

\section{Off-Policy Estimation and Confidence Interval} \label{sec:el}

We first review how empirical likelihood extends the above results, then present our results.

\subsection{Empirical Likelihood}

So far, we assumed that the random vector $(w,r)$ has finite support and that
data is iid. Empirical likelihood~\cite{owen2001empirical} allows us to
transfer the above results to settings where the support is infinite.  The
seminal work \cite{owen1988empirical} showed that the finite support assumption
is immaterial. This was later extended \cite{qin1994empirical} to prove that
estimating equations such as $\E[w]=1$ could be incorporated into the
estimation of $\E[wr]$.  As long as $\Cov(w,wr)\neq0$ Corollary~5 of
\cite{qin1994empirical} (also Theorem~3.5 of~\cite{owen2001empirical}) implies
that $-2(L(V(\pi))-\sum_n\log(Q^{\textrm{mle}}_{w_n,r_n})) \to \chi_{(1)}^2$ in
distribution as $n\to\infty$ without assuming finite support for $(w,r)$.
Asymptotic optimality results for empirical likelihood are established in
\cite{kitamura2001asymptotic}, but require different proof techniques from the
multinomial case \cite{hoeffding1965asymptotically}.

We now turn to the iid.\ assumption. In many practical setups data may have
been collected from various logging policies which makes the $w$'s non-iid.
Existing estimators, such as IPS, have no trouble handling such data. A key
insight is that all the information about the problem is captured in the
martingale estimating equation $m_n(V(\pi))=0$. The extension of empirical
likelihood to martingales is given by Dual Likelihood \cite{mykland1995dual}.
The reason for the name is that the functional of interest is the convex dual
of the empirical likelihood formulation subject to the martingale estimating
equation of interest.  In our case, we use dual variables $\tau$ and $\beta$
that correspond to the first and second component of $m_n(v)=0$ respectively.
As derived in appendix~\ref{app:lrdual} we get the dual likelihood
\begin{equation}
l_v(\beta,\tau) = \sum_n \log\left(1 + \beta (w_n - 1) + \tau (w_n r_n - v)\right)
\label{eqn:lrdual}
\end{equation}
  That derivation also reveals the constraint
set associated with a feasible primal solution,
\begin{equation}
\mathcal{C}=\{(\beta,\tau) | \forall w, r: 1 + \beta (w - 1) + \tau (w r - v) \geq 0\}.
\label{eqn:constraints}
\end{equation}
Despite the domains of $w$ and $r$ being potentially infinite,
we can express $\mathcal{C}$ using only 4 constraints as
$\mathcal{C} =\{(\beta,\tau) | \forall w \in \{ w_{\min}, w_{\max}\}, r \in \{ 0, 1 \}: 1 + \beta (w - 1) + \tau (w r - v) \geq 0 \}$.

This is also the convex dual of 
$\eqref{eq:primalprof}$ as iid and finite support data are 
just special cases of this framework. However, $L(v)$
and the corresponding $Q$ do not have a generative interpretation when
$w$'s are not iid.  Nevertheless, under very mild 
conditions~\cite{mykland1995dual}
the maximum of eq.~\eqref{eqn:lrdual} with $v=V(\pi)$
still has an asymptotic distribution that obeys a nonparametric analogue
to Wilks' theorem. Thus it functions similarly for hypothesis testing.
We will still refer to the support of $Q$ to provide intuition.

What is the set of alternative hypotheses considered when constructing 
hypothesis tests or CIs via a dual likelihood formulation? 
This is easier to understand 
in the primal, as the dual likelihood corresponds to a primal optimization over all
distributions $Q$ over $(w,r)$ which measure-theoretically dominate the empirical
distribution (i.e., place positive probability on each realized datum)
and satisfy the moment condition $\mathbb{E}_Q[w]=1$.
Although this includes distributions with unbounded support, the optima
are supported on the sample plus at most one more point as discussed in section~\ref{sec:constraint}.

\subsection{Off-Policy Estimation} \label{subsec:estimation}

We start by defining a (dual) analogue to the nonparametric maximum likelihood 
estimator (NPMLE) in the primal formulation for the iid case.
Consider the quantity
\begin{equation}
l^*_{\text{mle}}=\sup_{(\beta,0) \in \C} l_v(\beta, 0)
\end{equation}
which is obtained by setting $\tau=0$ (so the value of $v$ is 
immaterial) and optimizing over $\beta$. This quantity may seem
mysterious, but it corresponds to the NPMLE.
Indeed, $\tau=0$ means $\mathbb{E}_Q[wr]$ is free to take on 
any value, as in the primal maximum likelihood formulation.
We propose our estimator as any $v$ 
which obtains the maximum dual likelihood, i.e., any value
in the set
\newcommand{\estimateequation}{\left\{ v \biggr| \sup_{(\beta,\tau) \in \C} l_v(\beta, \tau) = l^*_{\text{mle}} \right\}}
\begin{equation}
\estimateequation.
\label{eqn:estimateset}
\end{equation}

In appendix~\ref{app:vhatmle} we prove there is an interval of maximizers of
the form
\begin{equation}
\hat V(\pi; \rho) = \rho + \frac{1}{N} \sum_n \frac{w_n (r_n - \rho)}{1 + \beta^* (w_n - 1)},
\label{eqn:vhatmle}
\end{equation}
where $\rho$ is any value in $[0, 1]$ and
$\beta^*$ maximizes
\begin{equation}
\sum_n \log\left(1 + \beta (w_n - 1)\right) \;\mathrm{s.t.}\; \forall w: 1 + \beta (w - 1) \geq 0.
\label{eqn:betastar}
\end{equation}
The constraints on $\beta^*$ are over all possible values of $w$,
not just the observed $w$.  However the constraints with $w=w_{\min}$ and $w=w_{\max}$ imply all other constraints.
We solve this 1-d convex problem via bisection to
accuracy $\epsilon$ in $O(N\log(\frac{1}{\epsilon}))$ time.
Note that $\beta=0$ is always feasible and it is optimal 
when $\sum_n w_n=N$. When $\beta^*=0$, \eqref{eqn:vhatmle} becomes $V^{\textrm{IPS}}$
for all values of $\rho$.

Eq.~\eqref{eqn:vhatmle} (and eq.~\eqref{eqn:cilbmle} in \ref{subsec:ci})
are valid in the martingale setting, i.e., for a sequence of
historical policies.  
Appendix~\ref{app:vhatmle} shows that when there exists an unobserved
extreme value of $w$, say $w_{ex}$, any associated primal solution  $Q^{\textrm{mle}}$ 
will assign some probability to a pair $(w_{ex},\rho)$.  Section~\ref{sec:constraint} discusses the beneficial implications
of this.  Once both $w_{\min}$, $w_{\max}$
are observed with any $r$, eq.~\eqref{eqn:vhatmle} becomes a point
estimate because $\sum_n w_n \left(1 + \beta^* (w_n - 1)\right)^{-1}=N$,
i.e., $\rho$ cancels out and $Q^{\textrm{mle}}$ only has support on the observed data.

The EMP estimator, based on empirical likelihood,
was proposed in \cite{kallus2019intrinsically}.  Specializing it
to a constant reward predictor for all $(x,a)$ we can
write both estimators in terms of $Q^{\textrm{mle}}$.
Eq.~\eqref{eqn:vhatmle} leads to $\hat{V}(\pi)=
(1-\sum_n Q_{w_n,r_n}^{\textrm{mle}} w_n) \rho + \sum_n Q_{w_n,r_n}^{\textrm{mle}} w_n r_n$ while
EMP is $\hat{V}_{\text{EMP}}(\pi)=\sum_n Q_{w_n,r_n}^{\textrm{mle}} w_n r_n/\sum_n
Q_{w_n,r_n}^{\textrm{mle}}$.  When $w_{\min}$ and $w_{\max}$ are observed, $\sum_n
Q_{w_n,r_n}^{\textrm{mle}}=\sum_n Q_{w_n,r_n}^{\textrm{mle}}w_n=1$ and the two estimators coincide.
Section~\ref{subsec:expestimation} empirically investigates their finite sample behavior.  

\subsubsection{Finite Sample Bias}
We show a finite-sample bound on the bias of an estimator, based upon
eq.~\eqref{eqn:vhatmle}, of the value difference $R(\pi) \doteq
V(\pi) - V(h)$ between $\pi$ and the logging policy.   We obtain our estimator for
$R(\pi)$ via $\E_{Q^\textrm{mle}}[wr]-\E_{Q^\textrm{mle}}[r]$ 
and using the primal-dual relationship for $Q^\textrm{mle}$ from appendix~\ref{app:lrdual}. In practical
applications $R(\pi)$ is the relevant quantity for deciding when to update
a production policy. The proof is in appendix~\ref{app:thmfinitebias}.
\newcommand{\textoftheorembias}{
	Let
	$\hat R(\pi) \doteq \frac{1}{N} \sum_n \frac{(w_n - 1) (r_n - \rho)}{1 + \beta^* (w_n - 1)}$
	with $\beta^*$ as in eq.~\eqref{eqn:betastar}, and let
	a.s. $\forall n: 0 \leq w_n \leq w_{\max}$ with $w_{\max} \geq 1$.  Then $$
	\left| \E\left[ \hat R(\pi) \right] - R(\pi) \right| \leq
	10 \sqrt{\frac{w_{\max}}{N}} + 16 \frac{w_{\max}}{N}
	$$
	where $R(\pi) \doteq V(\pi) - V(h)$ is the true policy value difference
	between $\pi$ and $\{ h_n \}_{n \in N}$.
}

\begin{theorem}
	\label{thm:finitebias}
	\textoftheorembias
\end{theorem}

The leading term in Theorem~\ref{thm:finitebias} is actually any $\omega\geq\E_{n}[(w_n-1)^2]$; $\omega=w_{\max}$
is a worst case.
This result indicates low bias for the estimator (leading terms
comparable to finite-sample variance); meanwhile inspection of
eq.~\eqref{eqn:vhatmle} (and \eqref{eqn:cilbmle}) indicate neither
can overflow the underlying range of reward.  This explains the excellent
mean square error performance observed in section~\ref{subsec:expestimation}.

\subsection{Off-Policy Confidence Interval} \label{subsec:ci}

We can use the dual likelihood of eq.~\eqref{eqn:lrdual} to construct an asymptotic confidence interval~\cite{mykland1995dual} in a manner completely analogous
to Wilks' theorem for the primal likelihood formulation
\newcommand{\eqnlrtest}{\left\{ v \biggr| \sup_{(\beta,\tau) \in \C} l_v(\beta, \tau) - l^*_{\text{mle}} \leq \frac{1}{2} \chi_{(1)}^{2,\alpha} \right\}}
\begin{equation}
\eqnlrtest,
\label{eqn:lrtest}
\end{equation}
where $\alpha$ is the desired nominal coverage and $\chi_{(1)}^{2,\alpha}$
is the $\alpha$-quantile of a $\chi$-square distribution with one degree
of freedom.  The asymptotic guarantee is that the coverage error of this
interval is $O(1/n)$.

In general applications of EL a bisection on $v$ is recommended for 
finding the boundaries of the CI: given
an interval $[\ell,u]$ check whether $v=(\ell+u)/2$ is in the set
given by \eqref{eqn:lrtest} and update $\ell$ or $u$.  This requires
$O(\log(1/\epsilon))$ calls to maximize \eqref{eqn:lrdual}. Here we derive
a more explicit form for the boundary points which is more insightful and
faster to compute (2 optimization calls).  In appendix~\ref{app:cimle}
we prove the lower bound of the CI is
\begin{equation}
v_{\text{lb}}(\pi) = \kappa^* \frac{1}{N} \sum_n \frac{w_n r_n}{\gamma^* + \beta^* w_n + w_n r_n},
\label{eqn:cilbmle}
\end{equation}
where $(\beta^*, \gamma^*, \kappa^*)$ are given by
\begin{equation*}
\sup_{\substack{\kappa \geq 0 \\ \beta, \gamma}} \sum_n \biggl(-\kappa \log \kappa + \kappa \bigl(-\phi + \log\left(\gamma + \beta w_n + w_n r_n \right) \bigr) \biggr)
\end{equation*}
subject to $\forall w: \gamma + \beta w \geq 0$, where $\phi =
\frac{1}{2N} \chi^{2,\alpha}_{(1)} - \frac{1}{N} l^*_{\text{mle}}$.

The constraints range over all possible values of $w$, but
$w_{\min}$ and $w_{\max}$ are the only relevant ones.  This is a
convex problem with 3 variables and 2 constraints that can
be solved to $\epsilon$-accuracy by the ellipsoid method (for example)
in $O(N\log(\frac{1}{\epsilon}))$ time.  The upper bound can be
obtained by transforming the rewards $r \leftarrow 1 - r$,
finding the lower bound, and then setting
$v_{\text{ub}} \leftarrow 1 - v_{\text{lb}}$.

In eq.~\eqref{eqn:cilbmle} we can have
$\kappa^* \sum_n w_n \left(\gamma^* + \beta^* (w_n - 1) + w_n r_n\right)^{-1} < N$
even after extreme values of $w$ have been observed.  This corresponds
to a primal solution which is placing ``extra'' probability on either
$(w_{\min}, 0)$ or $(w_{\max}, 0)$.  For 
example, this allows our lower bound to be $<1$ even if all observed rewards are $1$. Section~\ref{sec:constraint}
discusses the benefits of the additional primal support.

\subsection{The Importance of $\E[w] = 1$} \label{sec:constraint}

By inspection, the primal constraint $\E[w]=1$ can be infeasible for
a distribution only supported on the observed values if 1 is not
in the convex hull of observed importance weights.  Consequently
solutions to equations \eqref{eqn:vhatmle} and \eqref{eqn:cilbmle}
can correspond to distributions $Q$ in the primal formulation 
with support beyond the
observed values.  This is a known property of constrained empirical
likelihood~\cite{grendar2017multinomial}.

We precisely characterize the additional support as confined to a single
extreme point.  In appendix~\ref{app:vhatmle} we show the support of
the primal distribution associated with equation~\eqref{eqn:vhatmle}
is a subset of $\{ (w_n, r_n) \}_{n \leq N} \cup \{ (w_{ex}, \rho) \}$,
where $w_{ex} = w_{\min}$ if $\sum w_n \geq N$ and otherwise
$w_{ex} = w_{\max}$.  In appendix~\ref{app:cimle} we show the support
of the primal distribution associated with equation \eqref{eqn:cilbmle}
is $\{ (w_n, r_n) \}_{n \leq N} \cup \{ (w_{ex}, 0) \}$, where $w_{ex}$
is either $w_{\min}$ or $w_{\max}$.
We also point out that similar ideas have already been used for 
multi-armed bandits. For example, the empirical KL-UCB algorithm \cite{cappe2013kullback}
uses empirical likelihood to construct an upper confidence bound on each 
arm by considering distributions that can place additional mass on the 
largest possible reward.

Although the modification of the support from the observed data points 
seems modest, it greatly improves both the estimator and the CI.
Critically, both can produce values that are outside the convex hull
of the observations, but never overflow the possible range $[0, 1]$.
In contrast, empirical likelihood on the sample is constrained to
the convex hull of the observations; while empirical likelihood on
the bounded range without the $\E[w]=1$ constraint can produce value
estimates in the range $[0, w_{\max}]$.  Furthermore, we observe in
practice that our CIs approach nominal coverage values from above,
as in Figure~\ref{fig:cisynthetic}.  This is not typical behavior when
empirical likelihood is constrained to the sample.

Per Lemma 2.1 of ~\cite{owen2001empirical}, empirical likelihood can
only place $O(1/n)$ mass outside the sample. With our primal constraint
$\E[w]=1$ this mass is further limited to $O(1/w_{\max})$, and decreases
as the realized average importance weight approaches 1.  As seen in
Figure~\ref{fig:cisynthetic}, this can result in non-trivial CIs in the
regime $n < w_{\max}$ where other interval estimation techniques struggle.

\subsection{Offline Contextual Bandit Learning} \label{subsec:lblearn}

Here the goal is to learn a policy $\pi$ using a dataset
$\{ (x_n, a_n, p_n, r_n) \}_{n \in N}$, i.e., without
interacting with the system generating the data.
One strategy is to leverage a counterfactual estimator to reduce policy
learning to optimization~\cite{li2015counterfactual}, suggesting the
use of equation~\eqref{eqn:vhatmle} in the objective.

Alternatively we can instead optimize the lower bound of
equation~\eqref{eqn:cilbmle}.  In the iid.\ case optimizing the lower
bound corresponds to a variant of distributionally robust optimization.
The log-empirical likelihood for a distribution $Q$ is equivalent to
the KL divergence between the empirical distribution $\mathbb{1}/N$ and
$Q$. A likelihood maximizer $Q^{\text{mle}}$ attains the minimum such
KL divergence.  By optimizing the lower bound~\eqref{eqn:cilbmle} we
are performing distributionally robust optimization with uncertainty set
\[
\mathcal{Q(\pi)} = \left\{ Q \biggr| \E_Q[w(\pi)]=1,\ \text{KL}\left(\left.\left.\frac{\mathbb{1}}{N}\right|\right|Q\right) \leq B(\pi) \right\},
\]
where $B(\pi)=\text{KL}\left(\frac{\mathbb{1}}{N}||Q^{\text{mle}}(\pi)\right) + \frac{1}{2N}\chi_{(1)}^{2,\alpha}$ and we have made dependences on $\pi$ explicit. Given a set of policies $\Pi$ we can set up the game
\[
\max_{\pi \in \Pi} \min_{Q \in \mathcal{Q(\pi)}} \sum_{n} Q_{w(\pi)_n,r_n} w(\pi)_n r_n
\]
for finding the policy $\pi^*\in \Pi$ with the best reward lower bound.
For our experiments we use a heuristic alternating optimization strategy.
In one phase the policy is fixed and we find the optimal dual variables
associated with equation~\eqref{eqn:cilbmle}.  In the alternate phase
we find a policy with a better lower bound, i.e., a policy which improves
upon equation~\eqref{eqn:cilbmle} with dual variables held fixed.
Developing better methods for solving this game is deferred for future
work.

\section{Experiments}

The purpose of our experiments is to demonstrate the empirical
behavior of the proposed methods against other methods that
use the same information. Comparing against methods that
leverage or focus on reward predictors is therefore out of scope,
as reward predictors can help/hurt any method.  Our experiments
compare MSE of estimators (section~\ref{subsec:expestimation}),
confidence interval coverage and width (section~\ref{subsec:expci}),
and utility of lower bound optimization for off-policy learning
(section~\ref{subsec:explbo}).

Replication instructions are available in the supplement,
and replication software is available at
\url{http://github.com/pmineiro/elfcb}. All experiment
details are in the appendix.

\subsection{Off-Policy Estimation}\label{subsec:expestimation}

\begin{figure}[t]
	\centering
	\begin{minipage}[t]{.45\textwidth}
		\centering
		\strut\vspace*{-\baselineskip}\newline
		\includegraphics[width=\textwidth,keepaspectratio]{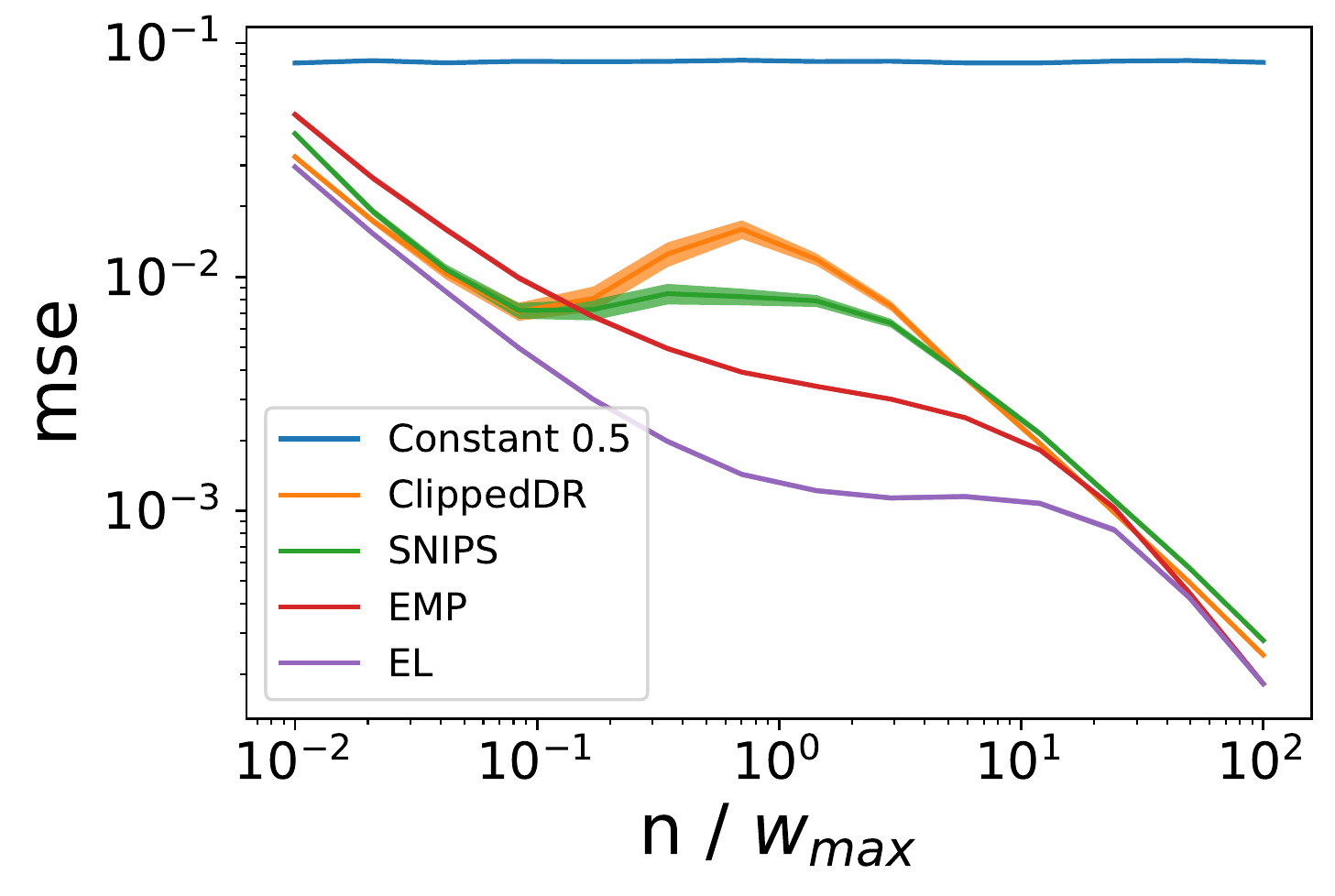}
		\captionof{figure}
		{Mean squared error of EL and other estimators on
			synthetic data. Asymptotics are similar while EL dominates in the
			small sample regime.  Line width is 4 times the standard error of the
			population mean.}
		\label{fig:mlesynthetic}
	\end{minipage}%
	\hfill
	\begin{minipage}[t]{.52\textwidth}
		\centering
		\strut\vspace*{-0.5 \baselineskip}\newline
    \begin{tabular}{lcccc}
	\toprule
	EL vs. & Exploration & Wins & Ties & Losses \\
	\midrule
	\multirow{3}{*}{IPS}
	& $\epsilon=0.05$ & 26 & 11 & 3 \\
	& bags=10 & 13 & 19 & 8 \\
	& cover=10 & 16 & 16 & 9 \\
	\midrule
	\multirow{3}{*}{SNIPS}
	& $\epsilon=0.05$ & 5 & 34 & 1 \\
	& bags=10 & 7 & 30 & 3 \\
	& cover=10 & 7 & 33 & 0 \\
	\midrule
	\multirow{3}{*}{EMP}
	& $\epsilon=0.05$ & 24 & 13 & 3 \\
	& bags=10 & 8 & 26 & 6 \\
	& cover=10 & 8 & 23 & 9 \\
	\bottomrule
\end{tabular}
  \captionof{table}
  {Off-policy evaluation results where $\epsilon=0.05$ is $\epsilon$-greedy exploration, bags=10 is bootstrap exploration with 10 replicas, and cover=10 is online cover~\cite{agarwal2014taming} with 10 policies.}
\label{tab:operesultspartial}
\end{minipage}
\end{figure}

\paragraph{Synthetic Data} We begin with
a synthetic example to build intuition. In appendix~\ref{app:opesynthexp} we detail how we sample $w=\pi/h$ and $r$ for each synthetic environment. Figure \ref{fig:mlesynthetic} shows the mean squared error (MSE)
over 10,000 environment samples for various estimators.  The best
constant predictor of 1/2 (``Constant'') has a MSE of 1/12, as
expected.  ClippedDR is the doubly robust estimator with the best
constant predictor of 1/2 clipped to the range $[0, 1]$, i.e.\ $\min(1, \max(0, \frac{1}{2} + \sum_n \frac{w_n}{N}
(r_n - 1/2)))$.
SNIPS is the self-normalized estimator IPS estimator. EMP is
the estimator of \cite{kallus2019intrinsically}. For EL,
we use $\rho = \frac{1}{2}$.  When a small number of large importance
weight events is expected in a realization, both ClippedDR and SNIPS
suffer due to their poor handling of the $\E[w] = 1$ constraint.
EMP is an improvement and EL is a further improvement.
Asymptotically all estimators are similar.

\paragraph{Realistic Data} We employ an
experimental protocol inspired by the operations of the Decision
Service~\cite{agarwal2016making}, an industrial contextual bandit
platform.  Details are in appendix~\ref{app:operdexp}.  Succinctly,
we use 40 classification datasets from OpenML~\cite{OpenML2013};
apply a supervised-to-bandit transform~\cite{dudik2011doubly};
and limit the datasets to 10,000 examples.
Each dataset is randomly split 20\%/60\%/20\%
into Initialize/Learn/Evaluate subsets, to learn $h$,
learn $\pi$, and evaluate $\pi$ respectively.  Learning is
via Vowpal Wabbit~\cite{langford2007vowpal} using various
exploration strategies, with default parameters
and $\pi$ initialized to $h$.

We compare the MSE of EL, IPS, SNIPS, and EMP using the
true value of $\pi$ on the evaluation set (available because the
underlying dataset is fully observed and $\pi(a|x)$ is known).  For each dataset we evaluate multiple times, each time resampling $a\sim h(\cdot |x)$.
Table \ref{tab:operesultspartial} shows the results of a paired
$t$-test with 60 trials per dataset and 95\% confidence level:
``tie'' indicates null result, and ``win'' or ``loss'' indicates
significantly better or worse.  The EL is overall superior to IPS and
SNIPS. It is similar to EMP except when the data comes from 0.05-greedy
exploration, where EL is better than EMP.

\begin{table*}[t]
\centering
\begin{tabular}{lcccccccc}
\toprule
\multirow{2}{*}{Exploration} & & \multicolumn{3}{c}{CI LB} & & \multicolumn{3}{c}{EL} \\
\cmidrule{3-5} \cmidrule{6-9}
 & & Wins & Ties & Losses & & Wins & Ties & Losses \\
\midrule
$\epsilon=0.05$ greedy & & 16 & 18 & 6 & & 11 & 26 & 3 \\
$\epsilon=0.1$ greedy & & 16 & 19 & 5 & & 13 & 24 & 3 \\
$\epsilon=0.25$ greedy & & 15 & 22 & 3 & & 3 & 34 & 3 \\
\midrule
bagging, 10 bags & & 21 & 18 & 1 & & 11 & 28 & 1 \\
bagging, 32 bags & & 4 & 26 & 10 & & 7 & 31 & 2 \\
\midrule
cover, 10 policies & & 18 & 21 & 1 & & 6 & 30 & 4 \\
cover, 32 policies & & 9 & 29 & 2 & & 6 & 34 & 0 \\
\bottomrule
\end{tabular}
\caption{Learning results.  ``CI LB'' uses equation~\eqref{eqn:cilbmle}; ``EL''
uses equation~\eqref{eqn:vhatmle}. ``EL'' serves as an ablation study,
on whether the improvement in ``CI LB'' is due to distributional robustness,
or the estimator itself.}
\label{tab:learnloggedresults}
\end{table*}

\subsection{Confidence Intervals} \label{subsec:expci}

\begin{wraptable}{r}{0.425 \linewidth}
	\vspace{-32pt}
	\centering
	\caption{Off-Policy Confidence Intervals} \label{tab:ciresults}
	\vspace{-2pt}
\begin{tabular}{lcc}
	\toprule
	Technique & Coverage & Width Ratio \\
	& (Average) & (Median) \\ \midrule
	EL & 0.975 & n/a  \\ \midrule
	Binomial & 0.996 & 2.89  \\ \midrule
	AG & 0.912 & 0.99  \\
	\bottomrule
\end{tabular}
	\vspace{-10pt}
\end{wraptable}

\paragraph{Synthetic Data} We use the same
synthetic $\epsilon$-greedy data as described above.  Figure
\ref{fig:cisynthetic} shows the mean width and empirical coverage
over 10,000 environment samples for various CIs at
95\% nominal coverage.  Binomial CI is the Clopper Pearson confidence
interval on the random variable $\frac{w}{w_{\max}}R$.
This is an excessively wide CI.  Asymptotic
Gaussian is the standard z-score CI around the
empirical mean and standard deviation motivated by the central limit
theorem. Intervals are narrow but typically violate nominal coverage.
The EL interval is narrow and obeys nominal coverage throughout
the entire range despite only having asymptotic guarantees.

Once again there is a qualitative change when the sample size
is comparable to the largest importance weight.  The Binomial CI
interval only begins to make progress at this point.  Meanwhile,
the asymptotic Gaussian interval widens as empirical variance
increases.

\paragraph{Realistic Data} We use the same datasets mentioned above,
but produce a 95\% confidence interval for off-policy evaluation
rather than the maximum likelihood estimate.  With 40 datasets and 60
evaluations per dataset, we have 2400 confidence intervals from which
we compute the coverage and the ratio of the width of the interval to
the EL in table~\ref{tab:ciresults}.  As expected from simulation,
the Binomial CI overcovers and has wider intervals.  EL widths are
comparable to asymptotic Gaussian (AG) on this data, but AG undercovers.
A 95\% binomial confidence interval on the coverage of AG is $[90.0\%,
92.3\%]$, indicating sufficient data to conclude undercoverage.

\subsection{Offline Contextual Bandit Learning}\label{subsec:explbo}

We use the same 40 datasets as above, but with a 20\%/20\%/60\%
Initialize/Learn/Evaluate split.  We made no effort to tune the confidence
level setting it to 95\% for all experiments. For optimizing the policy
parameters and the distribution dual variables, we alternate between
solving the dual problem with the policy fixed and then optimizing the
policy with the dual variables fixed.  To optimize the policy we do
a single pass over the data using Vowpal Wabbit as a black-box oracle
for learning, supplying different importance weights on each example
depending upon the dual variables.  We do 4 passes over the learning
set and update the dual variables before each pass.  Details are in
appendix~\ref{app:learnlogged}.

We compare the true value of $\pi$ on the evaluation set resulting
from learning with the different objectives. For each dataset
we learn multiple times, with different actions chosen by the
historical policy $h$.  Table \ref{tab:learnloggedresults} shows the
results of a paired $t$-test with 60 trials per dataset and 95\%
confidence level: ``tie'' indicates null result, and ``win'' or
``loss'' indicates significantly better or worse evaluation value
for the CI lower bound.  Using the CI lower bound overall yields
superior results.  Using the EL estimate also provides some lift
but is less effective than using the CI lower bound.

\section{Conclusions}

We presented a practical estimator and a CI for contextual bandits with correct
asymptotic coverage and empirically valid coverage for small
samples. To this end we used empirical likelihood techniques which
yielded computationally efficient and hyperparameter-free procedures
for estimation, CIs and learning. Empirically, our proposed CI is a
substantial improvement over existing methods and the learning algorithm
is a useful improvement against techniques that optimize the value of
a point estimate. Our methods offer the
largest advantage in regimes where existing methods struggle,
such as when the number of samples $N$ is of the same order as the 
largest possible importance weight.



\section*{Broader Impact}

Not applicable to this work.

\begin{ack}
We thank Adith Swaminathan and the anonymous reviewers for their 
valuable comments on earlier drafts on this work.
\end{ack}

\bibliographystyle{plain}
\bibliography{mlecb}

\newpage

\appendix
\isappendixtrue
\setcounter{theorem}{0}
\setcounter{lemma}{0}

\section{Derivation of Profile Likelihood} \label{app:lrdual}

For ease of exposition, we will start with a primal formulation and via
duality show equivalence with Dual Likelihood
\cite{mykland1995dual}
applied to the Dol\'{e}ans-Dade multiplicative martingale corresponding
to $m_n(v)$.

Starting from
\begin{equation*}
\sup_{Q \in \Delta}\left\{\left. \sum_n \log\left(Q_{w_n,r_n}\right) \right| \E_Q[w] = 1, \E_Q[w r] = v \right\}.
\end{equation*}
we form the Lagrangian dual
\begin{equation*}
\begin{split}
&\sup_{\beta,\gamma,\tau} \inf_{Q \succeq 0}\left\{\beta \left(-1 + \sum_{w,r} w Q_{w,r}\right) + \gamma \left(-1 + \sum_{w,r} Q_{w,r}\right) \right.\\
&\quad + \tau\left(-v + \sum_{w,r} w r Q_{w,r}\right) \left. - \sum_{w,r} c_{w,r} \log\left(Q_{w,r}\right) \right\},
\end{split}
\end{equation*}
where $c_{w,r} = \sum_n 1_{w=w_n,r=r_n}$.  Collecting terms
\begin{equation*}
\begin{split}
&\sup_{\beta,\gamma,\tau} \inf_{Q \succeq 0}\left\{-\beta - \gamma - \tau v \right. \\
&\quad \left. + \sum_{w,r} \left(\beta w + \gamma + \tau w r\right) Q_{w,r} - c_{w,r} \log\left(Q_{w,r}\right) \right\}.
\end{split}
\end{equation*}
Dual boundedness requires $\forall w, r: \beta w + \gamma + \tau w r
\geq 0$.  The infimum over $Q$ is separable yielding
\begin{equation*}
Q^*_{w,r} = \frac{c_{w,r}}{\beta w + \gamma + \tau w r},
\end{equation*}
if $c_{w,r} > 0$ or $\beta w + \gamma + \tau w r > 0$, otherwise the
contribution to the dual is zero.  Substituting
\begin{equation*}
\begin{split}
\sup_{\beta,\gamma,\tau} \biggl\{-\beta - \gamma - \tau v &\left. + \sum_n \log\left(\beta w_n + \gamma\right) \right. \\
&\biggr| \forall w: \beta w + \gamma + \tau w r \geq 0 \biggr\}
\end{split}
\end{equation*}
discarding constants.

Summing the KKT stationarity conditions yields $$
\begin{aligned}
\frac{c_{w,r}}{Q_{w,r}} &= \beta w + \gamma + \tau w r, \\
\sum_{w,r} c_{w,r} &= \beta \sum_{w,r} w Q_{w,r} + \gamma \sum_{w,r} Q_{w,r} + \tau \sum_{w,r} w r Q_{w,r}, \\
N &= \beta + \gamma + \tau v.
\end{aligned}
$$ Substituting, changing variables $\beta \leftarrow N \beta$ and
$\tau \leftarrow N \tau$, and discarding constants yields
\begin{equation*}
l_v(\beta,\tau) = \sum_n \log\left(1 + \beta (w_n - 1) + \tau (w_n r_n - v)\right).
\end{equation*}

\section{Derivation of Value Estimate} \label{app:vhatmle}
From equation~\eqref{eqn:estimateset},
\begin{equation}
\estimateequation,
\tag{\ref{eqn:estimateset}}
\end{equation}
we see any value estimate achieves the maximum dual likelihood value.
Applying the duality established in Appendix~\ref{app:lrdual} to
$l_v(\beta, 0)$ indicates all value estimates correspond to $v = \E_Q[w
r]$ where $Q$ achieves the primal maximum
\begin{equation*}
\begin{split}
\sup_{Q \in \Delta}\left\{\left. \sum_n \log\left(Q_{w_n,r_n}\right) \right| \E_{Q}\left[w\right] = 1 \right\}.
\end{split}
\end{equation*}
Forming the Lagrangian dual
\begin{equation*}
\begin{split}
\sup_{\beta,\gamma} \inf_{Q \succeq 0}&\left\{\beta \left(-1 + \sum_{w,r} w Q_{w,r}\right) + \gamma \left(-1 + \sum_{w,r} Q_{w,r}\right) \right.\\
&\left. - \sum_{w,r} c_{w,r} \log\left(Q_{w,r}\right) \right\},
\end{split}
\end{equation*}
where $c_{w,r} = \sum_n 1_{w=w_n,r=r_n}$.  Collecting terms
\begin{equation*}
\begin{split}
\sup_{\beta,\gamma} \inf_{Q \succeq 0}&\left\{-\beta - \gamma \right. \\
&\left. + \sum_{w,r} \left(\beta w + \gamma\right) Q_{w,r} - c_{w,r} \log\left(Q_{w,r}\right) \right\}.
\end{split}
\end{equation*}
Dual boundedness requires $\forall w: \beta w + \gamma \geq 0$.  The
infimum over $Q$ is separable yielding
\begin{equation*}
Q^*_{w,r} = \frac{c_{w,r}}{\beta w + \gamma},
\end{equation*}
if $c_{w,r} > 0$ or $\beta w + \gamma > 0$, otherwise the contribution
to the dual is zero.  Substituting
\begin{equation*}
\sup_{\beta,\gamma} \biggl\{-\beta - \gamma + \sum_n \log\left(\beta w_n + \gamma\right) \biggr| \forall w: \beta w + \gamma \geq 0 \biggr\}
\end{equation*}
discarding constants.

Summing the KKT stationarity conditions yields $$
\begin{aligned}
\frac{c_{w,r}}{Q_{w,r}} &= \beta w + \gamma, \\
\sum_{w,r} c_{w,r} &= \beta \sum_{w,r} w Q_{w,r} + \gamma \sum_{w,r} Q_{w,r}, \\
N &= \beta + \gamma.
\end{aligned}
$$ Substituting, changing variables $\beta \leftarrow N \beta$, and discarding
constants yields
\begin{equation*}
\begin{split}
\sup_{\beta} \left.\left\{\sum_n \log\left(\beta (w_n - 1) + 1\right)\right| \forall w: \beta (w - 1) + 1 \geq 0 \right\}.
\end{split}
\end{equation*}

If $\beta^* = 0$ then $Q^*$ is supported only on the sample due to $1
= \E_Q[1]$.  Otherwise, $Q^*$ is entirely supported on the sample except
where $1 + \beta^* (w - 1) \geq 0$ is satisfied with equality.  This can
only be at the smallest or largest possible value of $w$ depending upon
the sign of $\beta^*$; call this $w_{ex}$.  Any $r$ is equally likely
at this point; call it $\rho$.

Equation~\eqref{eqn:vhatmle} follows via $$
\begin{aligned}
\hat V(\pi) &= \sum_{w,r} w Q_{w,r} r \\
&= \sum_n w_n Q_{w_n,r_n} r_n + w_{ex} Q_{w_{ex},\rho} \rho \\
&= \sum_n w_n Q_{w_n,r_n} r_n + \left(1 - \sum_n w_n Q_{w_n, r_n}\right) \rho \\
&= \rho + \sum_n w_n Q_{w_n,r_n} (r_n - \rho) \\
&= \rho + \frac{1}{N} \sum_n \frac{w_n (r_n - \rho)}{1 + \beta^* (w_n - 1)}
\end{aligned}
$$ where the first line is by definition, the third by $1 = \E_Q[w]$,
and the fifth line by the primal-dual relationship.

\section{Derivation of Lower Bound} \label{app:cimle}
The lower bound is the infimum of the value set defined by equation~\eqref{eqn:lrtest},
\begin{equation}
\eqnlrtest.
\tag{\ref{eqn:lrtest}}
\end{equation}
Applying the duality established in Appendix~\ref{app:lrdual}
we get the equivalent primal formulation
$$
\inf_{Q \in \Delta} \left\{ \E_Q[w r] \Bigr| \E_Q[w] = 1, \sum_n \log(Q_{w_n,r_n}) \geq \phi \right\}
$$ where $\phi = \sum_n \log(Q^{\text{mle}}_{w_n, r_n}) -\frac{1}{2} \chi^{2,\alpha}_{(1)}$.  A Lagrangian dual is
\begin{equation*}
\begin{split}
&\sup_{\substack{\kappa \geq 0\\\beta,\gamma}} \inf_{Q \succeq 0}\left\{\beta \left(-1 + \sum_{w,r} w Q_{w,r}\right) + \gamma \left(-1 + \sum_{w,r} Q_{w,r}\right) \right.\\
&\quad + \left. \kappa\left(\phi - \sum_{w,r} c_{w,r} \log\left(Q_{w,r}\right) \right) + \sum_{w,r} w r Q_{w,r} \right\},
\end{split}
\end{equation*}
where $c_{w,r} = \sum_n 1_{w=w_n,r=r_n}$.  Collecting terms
\begin{equation*}
\begin{split}
\sup_{\substack{\kappa \geq 0\\\beta,\gamma}} \inf_{Q \succeq 0}&\left\{-\beta - \gamma + \kappa \phi \right. \\
&\left. + \sum_{w,r} \left(\beta w + \gamma + w r\right) Q_{w,r} - \kappa c_{w,r} \log\left(Q_{w,r}\right) \right\}.
\end{split}
\end{equation*}
Dual boundedness requires $\forall w,r: \beta w + \gamma + w r > 0 \lor \left( \beta w + \gamma + w r = 0 \land c_{w,r} = 0 \right)$.  The
infimum over $Q$ is separable yielding
\begin{equation*}
Q^*_{w,r} = \kappa \frac{c_{w,r}}{\beta w + \gamma + w r},
\end{equation*}
if $c_{w,r} > 0$ or $\beta w + \gamma + w r > 0$, otherwise the
contribution to the dual is zero.  Substituting and changing
variables $\phi \leftarrow \frac{\phi - 1}{N}$ yields
\begin{equation*}
\sup_{\substack{\kappa \geq 0 \\ \beta, \gamma}} -\beta -\gamma + \sum_n \biggl(-\kappa \log \kappa + \kappa \bigl(\phi + 1 + \log\left(\gamma + \beta w_n + w_n r_n \right) \bigr) \biggr)
\end{equation*}
discarding constants.

$Q^*$ is supported on the sample except where $\beta w + \gamma + w r
\geq 0$ is satisfied with equality.  Because $w r \geq 0$, this implies
equality can only happen at $w r = 0$ otherwise other violations occur.
Thus all constraints are implied by $\forall w \in \{w_{\min}, w_{\max}\}:
\beta w + \gamma \geq 0$.  Denote $\Xi$ to be the set of $(w, r)$
pairs where equality occurs.

Equation~\eqref{eqn:cilbmle} follows via $$
\begin{aligned}
v_{\text{lb}}(\pi) &= \sum_{w,r} Q_{w,r} w r \\
&= \frac{1}{N} \sum_n Q_{w_n,r_n} w_n r_n + \sum_{(w, r) \in \Xi} Q_{w, r} w r \\
&= \frac{1}{N} \sum_n w_n Q_{w_n,r_n} r_n & (\forall (w, r) \in \Xi: w r = 0) \\
&= \kappa^* \frac{1}{N} \sum_n \frac{w_n r_n}{\gamma^* + \beta^* w_n + w_n r_n},
\end{aligned}
$$ where the first line is by definition,
and the fourth line by the primal-dual relationship.

\section{Proof of Theorem \ref{thm:finitebias}} \label{app:thmfinitebias}

\begin{lemma}
Let $\beta^*$ solve
\begin{equation*}
\sup_\beta\left\{ \left.\sum_n \log\left(1 + \beta (w_n - 1)\right) \right| \forall w: 1 + \beta (w - 1) \geq 0 \right\}.
\end{equation*}
Then $$
\begin{aligned}
\left| \beta^* \right| \sum_n \frac{(w_n - 1)^2}{1 + \beta^* (w_n - 1)} &\leq \left| \sum_n (w_n - 1) \right|.
\end{aligned}
$$
\begin{proof}
For the unconstrained maximizer, $$
\begin{aligned}
0 &= \sum_n \frac{w_n - 1}{1 + \beta^* (w_n - 1)} \\
&= \sum_n (w_n - 1) \left( 1 - \frac{\beta^* (w_n - 1)}{1 + \beta^* (w_n - 1)} \right), \\
\end{aligned}
$$
$$
\begin{aligned}
\beta^* \sum_n \frac{(w_n - 1)^2}{1 + \beta^* (w_n - 1)} &= \sum_n (w_n - 1), \\
\left| \beta^* \right| \sum_n \frac{(w_n - 1)^2}{1 + \beta^* (w_n - 1)} &= \left| \sum_n (w_n - 1) \right|.
\end{aligned}
$$ For the constrained maximizer, first note the sign of $\beta^*$ is the sign of $\sum_n (w_n - 1)$ because $\beta = 0$ is feasible and $$
\begin{aligned}
\left. \frac{\partial}{\partial \beta} \sum_n \log\left(1 + \beta (w_n - 1)\right) \right|_{\beta=0} &= \sum_n (w_n - 1).
\end{aligned}
$$
If the constrained maximizer is positive than $$
\begin{aligned}
0 &< \left. \frac{\partial}{\partial \beta} \sum_n \log\left(1 + \beta (w_n - 1)\right)\right|_{\beta=\beta^*} \\
&= \sum_n \frac{w_n - 1}{1 + \beta^* (w_n - 1) } \\
&= \sum_n (w_n - 1) \left( 1 - \frac{\beta^* (w_n - 1)}{1 + \beta^* (w_n - 1)} \right), \\
\end{aligned}
$$
$$
\begin{aligned}
\beta^* \sum_n \frac{(w_n - 1)^2}{1 + \beta^* (w_n - 1)} &< \sum_n (w_n - 1), \\
\left| \beta^* \right| \sum_n \frac{(w_n - 1)^2}{1 + \beta^* (w_n - 1)} &< \left| \sum_n (w_n - 1) \right|.
\end{aligned}
$$ If the constrained maximizer is negative than $$
\begin{aligned}
0 &> \left. \frac{\partial}{\partial \beta} \sum_n \log\left(1 + \beta (w_n - 1)\right)\right|_{\beta=\beta^*} \\
&= \sum_n \frac{w_n - 1}{1 + \beta^* (w_n - 1)} \\
&= \sum_n (w_n - 1) \left( 1 - \frac{\beta^* (w_n - 1)}{1 + \beta^* (w_n - 1)} \right), \\
\end{aligned}
$$
$$
\begin{aligned}
\beta^* \sum_n \frac{(w_n - 1)^2}{1 + \beta^* (w_n - 1)} &> \sum_n (w_n - 1), \\
\left| \beta^* \right| \sum_n \frac{(w_n - 1)^2}{1 + \beta^* (w_n - 1)} &< \left| \sum_n (w_n - 1) \right|.
\end{aligned}
$$.
\end{proof}
\end{lemma}

\begin{lemma}
Let $\{ \sum_{k \leq n} (w_k - 1) \}_{n \in N}$ be a martingale sequence
adapted to the filtration $\{ \mathcal{F}_n \}_{n \in N}$ where
a.s. $\forall n: 0 \leq w_n \leq w_{\max}$ with $w_{\max} \geq 1$.  Then $$
\E\left[\frac{1}{N} \left| \sum_{n \leq N} (w_n - 1) \right| \right] \leq 5 \sqrt{\frac{y}{N}} + 8 \frac{w_{\max}}{N},
$$
where a.s. $$
y \geq \frac{1}{N} \sum_{n \leq N} \E\left[(w_n - 1)^2 | \mathcal{F}_n\right].
$$
\begin{proof}
Freedman's inequality indicates $$
\mathrm{Pr}\left( |M_N| \geq x, \langle M \rangle_N \leq y\right) \leq 2 \exp\left(-\frac{x^2}{2 (y + w_{\max} x)}\right),
$$ where $M_N \doteq \sum_n \Delta M_n$, $\Delta M_n \doteq w_n - 1$,
$\langle M \rangle_N \doteq \sum_{n \leq N} \E\left[ \Delta M_n^2 | \mathcal{F}_{n-1}\right]$.  Let $\mathcal{E}$ denote the event $\langle M \rangle_N \leq y$.  Then
$$
\begin{aligned}
\E\left[ \left| M_N \right| 1_{\mathcal{E}} \right] &=
\int_0^\infty dx\ \mathrm{Pr}\left( \left| M_N \right| \geq x, \mathcal{E} \right).
\end{aligned}
$$  We do the integration in pieces.  For $x \geq \frac{y}{w_{\max}}$, we have
$$
\begin{aligned}
& \int_{\frac{y}{w_{\max}}}^\infty dx\ \mathrm{Pr}\left( \left| M_N \right| \geq x, \mathcal{E} \right) \\
&= \int_{\frac{y}{w_{\max}}}^\infty dx\ 2 \exp\left(-\frac{x^2}{2 (y + w_{\max} x)}\right) \\
&\leq \int_{\frac{y}{w_{\max}}}^\infty dx\ 2 \exp\left(-\frac{x}{4 w_{\max}}\right) \\
&= 8 w_{\max} \exp\left(-\frac{y}{4 w_{\max}^2}\right) \\
&\leq 8 w_{\max}.
\end{aligned}
$$  Therefore
$$
\begin{aligned}
&\E\left[ \left| M_N \right| 1_{\mathcal{E}} \right] \\
&\leq 8 w_{\max} + \int_0^{\frac{y}{w_{\max}}} 2 \exp\left(-\frac{x^2}{2 (y + w_{\max} x)}\right) \\
&\leq 8 w_{\max} + a + \int_a^{\frac{y}{w_{\max}}} 2 \exp\left(-\frac{x^2}{2 (y + w_{\max} x)}\right) \\
&\leq 8 w_{\max} + a + \int_a^{\frac{y}{w_{\max}}} 2 \exp\left(-\frac{x^2}{4 y}\right) \\
&\leq 8 w_{\max} + a + 2 \sqrt{\pi} \sqrt{y} \left(1 - \erf\left(\frac{a}{2 \sqrt{y}}\right)\right) \\
&\leq 8 w_{\max} + 2 \sqrt{y} \left(\sqrt{\pi} \left(1 + \erf\left(\sqrt{\log(2)}\right) \right) - \sqrt{\log(2)}\right) \\
&\leq 5 \sqrt{y} + 8 w_{\max}.
\end{aligned}
$$
Dividing by $N$ completes the proof.
\end{proof}
\end{lemma}

\begin{theorem}
\textoftheorembias
\begin{proof}
Consider the random variable $$
\begin{aligned}
\Delta \hat{R}(\pi) &= \hat R(\pi) - \frac{1}{N} \sum_n (w_n - 1) (r_n - \rho) \\
&= \frac{1}{N} \sum_n \frac{\beta^* (w_n - 1)^2}{1 + \beta^* (w_n - 1)} (r_n - \rho).
\end{aligned}
$$ $\Delta \hat{R}(\pi)$ is the difference of $\hat R(\pi)$ and an unbiased
estimator, therefore its expectation is the bias of $\hat R(\pi)$. $$
\begin{aligned}
&\left| \E\left[ \Delta \hat{R}(\pi) \right] \right| \\
&\leq \E\left[ \left| \Delta \hat{R}(\pi) \right| \right] \\
&\leq 2 \E\left[ \frac{1}{N} \left|\beta^*\right| \sum_n \frac{(w_n - 1)^2}{1 + \beta^* (w_n - 1)} \right] \\
&\leq 2 \E\left[ \frac{1}{N} \left| \sum_n (w_n - 1) \right| \right] \\
&\leq 10 \sqrt{\frac{y}{N}} + 16 \frac{w_{\max}}{N}.
\end{aligned}
$$ Finally we can bound $y$ via
$\E[(w_n - 1)^2 | \mathcal{F}_{n-1}] \leq \E[w_n^2 | \mathcal{F}_{n-1}] \leq w_{\max} \E[w_n | \mathcal{F}_{n-1}] \leq w_{\max}.$
\end{proof}
\end{theorem}

\section{Off-Policy Evaluation, Synthetic Data} \label{app:opesynthexp}
First, an environment
is sampled.  For all environments, the historical logging policy is
$\epsilon$-greedy with possible importance weights $(0, 2, 1000)$.
We choose $\pi$ to induce the maximum entropy distribution over
importance weights consistent with $\E[w^2] = 100$.
Rewards are binary with the conditional distribution of reward
varying per environment draw such that the value of $\pi$ is
uniformly distributed on $[0, 1]$.  Once an environment is drawn a
set of examples is sampled from that environment, and the squared
error of the value estimate is computed.

\section{Off-Policy Evaluation, Realistic Data} \label{app:operdexp}

We use the following 40 datasets from OpenML~\cite{OpenML2013}
identified by their OpenML dataset id: 1216, 1217, 1218, 1233, 1235,
1236, 1237, 1238, 1241, 1242, 1412, 1413, 1441, 1442, 1443, 1444,
1449, 1451, 1453, 1454, 1455, 1457, 1459, 1460, 1464, 1467, 1470,
1471, 1472, 1473, 1475, 1481, 1482, 1483, 1486, 1487, 1488, 1489,
1496, 1498.  For each dataset we convert to Vowpal Wabbit format,
shuffle the dataset, and utilize up to the first 10,000 examples
as data.  We utilize a 20\%/60\%/20\% Initialize/Learn/Evaluate
split sequentially by line number.  Note the shuffle and split
is done only once per dataset.  We create a historical policy $h$
using on-policy learning on the Initialize dataset, and then learn
a new policy $\pi$ on the Learn dataset using off-policy learning
with data drawn from $h$.  These Initialize and Learn steps are
done once per dataset.  Only the off-policy evaluation step is
done multiple times per dataset, and the random variations are due
to the different actions selected by $h$ over the Evaluate set.
For each evaluation, we compute the squared error of the different
predictors, i.e., the squared difference between the off-policy
value estimate and the true value of $\pi$.  Note the true value
of $\pi$ can be computed (and is independent of the choices of $h$
on the evaluation set) because the underlying datasets are fully
observed. Using the squared error as the random variable, we apply
a paired $t$-test between EL and the other predictors to determine
win, loss, or tie for each dataset.  We use default settings for
Vowpal Wabbit except for the choice of exploration strategy.

\section{Learning from Logged Bandit Feedback}
\label{app:learnlogged}

We first utilize the same 40 datasets as above, but with a 20\%/20\%/60\%
Initialize/Learn/Evaluate split.  The Initialize step is done once per
dataset, then the Learn and Evaluate steps are done multiple times
per dataset.  Note the Evaluate step here is using the true value
of $\pi$, i.e., is deterministic and independent of $h$ given $\pi$.
Using the evaluation score as the random variable, we apply a paired
$t$-test between MLE and the other predictors to determine win, loss,
or tie for each dataset.  We use Vowpal Wabbit in IPS learning mode
with default settings, and do 4 passes over the data.  At the beginning
of each pass, we optimize the dual variables holding the policy fixed,
then use the resulting dual variables during the learning pass to
compute importance weights.

\makeatletter
\if@submission
\makeatother
\relax
\else
\makeatother
\section{Cressie-Read Divergence Results}

We describe variants of the estimator and confidence interval
utilizing the Cressie-Read power divergence, which takes the form
\begin{equation*}
\text{CR}(\lambda) = \frac{2}{\lambda (\lambda + 1)} \sum_n \left( \left( N Q_{w_n,r_n} \right)^{-\lambda} - 1\right)
\end{equation*}
with parameter $\lambda$.  The choice $\lambda = -2$ is of practical
interest because it yields closed-form solutions driven by sufficient
statistics that are easily maintained online.

\subsection{Estimator}
The primal formulation for the estimator is
\begin{equation*}
\sup_{Q \in \Delta}\left\{\left. \sum_n \left( \left(N Q_{w_n,r_n}\right)^2 - 1 \right)\right| \E_Q[w] = 1 \right\}.
\end{equation*}
When optimizing over all distributions this can result in all the
mass placed outside the sample, so we constrain the distributions to
be supported on the empirical support plus an additional importance
weight $w_{\text{undata}}$, with arbitrary associated reward $\rho$,
corresponding to where the KL divergence places additional support:
\begin{equation*}
w_{\text{undata}} =
\begin{cases}
w_{\min} & \frac{1}{N} \sum_n w_n \geq 1 \\
w_{\max} & \text{otherwise}
\end{cases}.
\end{equation*}
This results in closed form solution
\begin{equation*}
Q_{w,r} = -\frac{\gamma^* + \beta^* w}{2 (N + 1)},
\end{equation*}
where $$
\begin{aligned}
\left( \begin{matrix} \gamma^* \\ \beta^* \end{matrix} \right) &= \left( \begin{matrix} \frac{b}{b - a^2} \\ -\frac{a}{b - a^2} \end{matrix} \right), \\
a &\doteq \frac{1}{N+1} \sum_{n \cup \{\text{undata}\}} (w_n - 1), \\
b &\doteq \frac{1}{N+1} \sum_{n \cup \{\text{undata}\}} (w_n - 1)^2.
\end{aligned}
$$ The resulting value estimate interval is $$
\hat V(\pi) = \rho + \frac{1}{N} \sum_n \left(\left(\frac{N}{1+N}\right) \gamma^* w_n + \left(\frac{N}{1+N}\right) \beta^* (w_n - 1)^2 \right) (r_n - \rho)
$$ where $\rho \in [0, 1]$.
Sufficient statistics for the estimator are $N$ and the (unaugmented support)
empirical sums of $w$ and $w^2$.

\subsection{Confidence Interval}
The primal formulation for the lower bound is
$$
\inf_{Q \in \Delta} \left\{ \E_Q[w r] \Bigr| \E_Q[w] = 1, \sum_n \left(N Q_{w_n,r_n}\right)^2 - 1 \geq \phi \right\}
$$
where $\phi = \frac{1}{2} \chi^{2,\alpha}_{(1)} - \left( \sum_n \left(N Q^{\text{mle}}_{w_n, r_n}\right)^2 - 1 \right)$.

When optimizing over all distributions this can result in all the mass
placed outside the sample, so we constrain the distributions to be
supported on the empirical support plus an additional importance weight
and reward pair.  We consider both extreme points
$\{ (w, 0) | w \in \{ w_{\min}, w_{\max} \} \}$
corresponding to where the KL divergence might place additional support,
and use the minimum value as the lower bound.  This results in a
closed-form solution
\begin{equation*}
Q_{w,r} = -\frac{\gamma^* + \beta^* w + w r}{(N+1) \kappa^*}, \\
\end{equation*}
where $$
\begin{aligned}
\left(\begin{matrix} \gamma^* \\ \beta^* \end{matrix}\right) &= \kappa^* \vec{a} + \vec{b}, \\
\vec{a} &\doteq \frac{1}{\overline{w^2} - \overline{w}^2} \left(\begin{matrix} -\overline{w^2} & \overline{w} \\ \overline{w} & -1 \end{matrix}\right) \vec{1}, \\
\vec{b} &\doteq  \frac{1}{\overline{w^2} - \overline{w}^2} \left(\begin{matrix} -\overline{w^2} & \overline{w} \\ \overline{w} & -1 \end{matrix}\right) \left(\begin{matrix} \overline{w r} \\ \overline{w^2 r} \end{matrix}\right),  \\
x &\doteq \overline{w r} + \frac{\left(1 - \overline{w}\right) \left(\overline{w^2 r} - \overline{w}\ \overline{w r} \right)}{\overline{w^2} - \overline{w}^2}, \\
y &\doteq \frac{\left(\overline{w^2 r} - \overline{w}\ \overline{w r}\right)^2}{\overline{w^2} - \overline{w}^2} - \left( \overline{w^2 r^2} - \overline{wr}^2 \right), \\
z &\doteq \phi + \frac{\left(1 - \overline{w}\right)^2}{2 \left(\overline{w^2} - (\overline{w})^2\right)}, \\
\kappa^* &= \sqrt{\frac{y}{2 z}},
\end{aligned}
$$
where $\overline{(\cdot)}$ denotes empirical mean including augmented support.
The resulting lower bound is $v_{\text{lb}}(\pi) = x - \sqrt{2 y z}$.

Sufficient statistics for the lower bound are $N$ and the (unaugmented support)
empirical sums of $w$, $w^2$, $w r$, $w^2 r$, and $w^2 r^2$.
\fi

\end{document}